\pdfoutput=1

\documentclass[11pt]{article}

\usepackage[final]{acl}

\usepackage{times}
\usepackage{latexsym}

\usepackage{amsmath} 
\usepackage{amssymb} 
\usepackage{listings}
\usepackage{lipsum}

\usepackage[T1]{fontenc}

\usepackage[utf8]{inputenc}

\usepackage{microtype}

\usepackage{inconsolata}

\usepackage{graphicx}

\title{SedarEval: Automated Evaluation using Self-Adaptive Rubrics  }

\usepackage{diagbox}
\usepackage{booktabs}
\usepackage{tabularx}
\usepackage{xcolor}
\usepackage{subcaption}
\usepackage{floatrow}
\usepackage{tikz}
\usepackage{wrapfig}
\usepackage{soul}
\usepackage{pifont}
\usepackage{multirow}

\usepackage[most]{tcolorbox}
\newtcolorbox{myquote}[1][]{%
    colback=black!3,
    colframe=black!3,
    notitle,
    sharp corners,
    borderline west={2pt}{0pt}{blue!80!black},
    enhanced,
    breakable,
}

\usepackage{algorithm}
\usepackage{algpseudocode}

\author{
 \textbf{Zhiyuan Fan\textsuperscript{1}\thanks{This work was done during Zhiyuan Fan’s internship at Xiaohongshu Inc.}},
 \textbf{Weinong Wang\textsuperscript{2}\thanks{Corresponding author. weinong.wang@hotmail.com}},
 \textbf{Xing Wu\textsuperscript{3}},
 \textbf{Debing Zhang\textsuperscript{2}}
\\
\\
 \textsuperscript{1}Hong Kong University of Science and Technology,
 \textsuperscript{2}Xiaohongshu Inc,
 \\
 \textsuperscript{3}Institute of Information Engineering, Chinese Academy of Sciences \\
}

\begin{document}
\maketitle


\begin{abstract}
The evaluation paradigm of LLM-as-judge gains popularity due to its significant reduction in human labor and time costs. This approach utilizes one or more large language models (LLMs) to assess the quality of outputs from other LLMs. However, existing methods rely on generic scoring rubrics that fail to consider the specificities of each question and its problem-solving process, compromising precision and stability in assessments. Inspired by human examination scoring processes, we propose a new evaluation paradigm based on self-adaptive rubrics. Specifically, we create detailed scoring rubrics for each question, capturing the primary and secondary criteria in a structured format of scoring and deduction points that mimic a human evaluator's analytical process. 
 %
 Building on this paradigm, we further develop a novel benchmark called SedarEval, which covers a range of domains including long-tail knowledge, mathematics, coding, and logical reasoning. SedarEval consists of 1,000 meticulously crafted questions, each with its own self-adaptive rubric. To further streamline the evaluation, we train a specialized evaluator language model (evaluator LM) to supplant human graders. Using the same training data, our evaluator LM achieves a higher concordance rate with human grading results than other paradigms, including GPT-4, highlighting the superiority and efficiency of our approach. We release our dataset at \url{https://github.com/wwn1233/sedareval}.



\end{abstract}

\section{Introduction}


The rapid advancements in large language models (LLMs) have led to their widespread use \cite{openai2024gpt4,team2023gemini,anthropic2024claude,bai2023qwen}. However, assessing these models in open-ended question-answering scenarios poses a significant challenge. Automated metric-based evaluations offer speed and convenience but often fall short due to the diversity of ground truth \cite{schluter-2017-limits,reiter-2018-structured,montahaei2019jointly,freitag-etal-2020-bleu}. In contrast, human-based evaluations provide reliable assessments but require substantial resources. 


To bridge the gap, the LLM-as-a-judge paradigm attempts to strike a balance between automated and human evaluation. Prominent examples of this approach include MT-bench \cite{zheng2024judging} and Arena \cite{chiang2024chatbot}, which leverage proprietary models to evaluate individual or comparative model responses. These benchmarks use pre-defined principles, such as the 3H principle (human-like, helpful, harmonious), to determine responses that align best with realistic human preferences. The widespread use of GPT-4 \cite{openai2024gpt4} as an evaluator in these studies presents challenges, including high costs for research institutions and potential data leaks.

Some studies \cite{zhu2023judgelm,li2023generative,wang2024pandalm,kim2024prometheus,kim2024prometheus2} propose using open-source pretrained models \cite{touvron2023llama,bai2023qwen,zeng2022glm} to train specialized evaluator LMs, offering a more cost-effective and secure solution. However, these methods typically use a uniform, question-agnostic rubric to guide the scoring process, overlooking the unique characteristics of each question. Each question has different emphases, with primary and secondary scoring points. A general rubric applies uniform criteria, failing to accurately reflect human preferences.


\begin{figure*}[t]
    \centering
    \includegraphics[width=\linewidth]{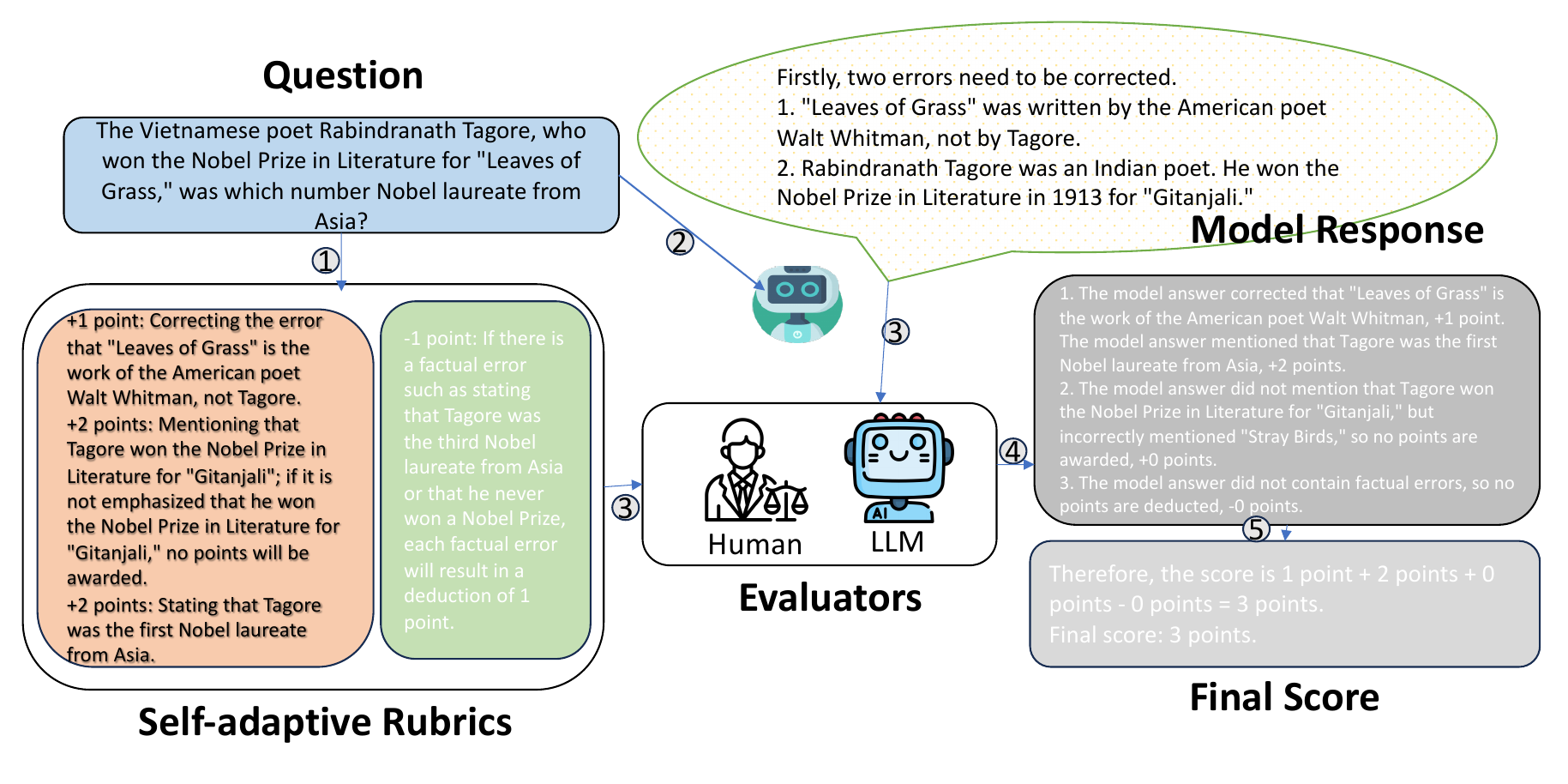}
    \caption{Automated evaluation pipeline using self-adaptive rubrics.This pipeline dynamically adjusts the evaluation rubric based on the input question, resulting in a scoring process that aligns more closely with human evaluators.}
    \label{fig:pipeline}
\end{figure*}

To adaptively align the scoring process with human judgment, we propose a novel evaluation paradigm based on self-adaptive rubrics. Unlike coarse-grained general rubrics, we provide fine-grained rubrics for each task, detailing specific scoring and penalty points with primary and secondary information. By analyzing focus points, we assign different values to each point. Additionally, we introduce penalty points to penalize models for generating rejected responses. The scoring process considers both preferred and rejected perspectives. The inconsistent coverage of positive and penalty points ensures a more refined constraint on the scoring process. These detailed scoring trajectories simplify the evaluation process to an instruction-following task, reducing dependency on a judge model's internal knowledge and skills, leading to more accurate and stable assessments. Building on this paradigm, we construct a new benchmark called SedarEval,
presenting a taxonomy comprising 8 major categories, with a dataset of 1000 queries, that fully aligns with realistic scenarios.

We further conduct ablation experiments on each component of the LLM-as-a-judge paradigm to investigate training a specialized LLM for scoring, revealing their respective importance. We analyze whether LLMs can correctly evaluate questions they can correctly answer and find that insufficient diversity in existing SFT data and a lack of evaluation-format data limit model performance. We also propose human-AI consistency to ensure evaluator LLMs maintain alignment with human preferences while leveraging their chain of thought capability to improve evaluation performance. Based on these findings, we develop a specialized evaluator LLM tailored to the benchmark for automated scoring. This model surpasses GPT-4 in model-level and question-level Pearson correlation, GSB, and ACC metrics, demonstrating higher consistency with human judgment. Experimental results validate the effectiveness and efficiency of our proposed paradigm.

Our contributions are summarized as follows:
\begin{enumerate}

\item We propose a novel evaluation paradigm using self-adaptive rubrics for each question, offering granular guidance and closely aligning the scoring process with human evaluation.

\item We develop a high-quality benchmark called SedarEval, featuring 1,000 meticulously crafted questions with detailed rubrics, and conduct manual evaluations on 20 LLMs.

\item We analyze the training of evaluator LMs, highlight existing methods' shortcomings, and use the self-adaptive rubrics paradigm to train an evaluator LM that surpasses GPT-4 in agreement with human evaluations.
\end{enumerate}


\section{Related Work}
\textbf{Benchmark LLMs Capabilities.} With the rapid advancement of LLMs~\cite{openai2024gpt4,team2023gemini,anthropic2024claude}, it has become a substantial challenge to benchmark their broad capabilities reliably. NLU-style tasks~\cite{hendrycks2020measuring, huang2024c, srivastava2022beyond, zhong2023agieval}, such as multi-choice QA, employ general-exam questions from various domains to assess a model's knowledge and comprehension abilities. However, their real-world usage is limited due to misalignment with human preferences. Recently, reference-free benchmarks~\cite{alpaca_eval,chiang2023vicuna,zheng2024judging,ye2023flask} have been proposed to evaluate texts' quality in a generative setting directly. Unlike previous datasets, our benchmark provides a comprehensive and stable model assessment with its diverse test cases and broad label distribution.

\setlength{\parindent}{0pt}\textbf{Automatic NLG Evaluation.} It's notably challenging to evaluate the quality of generated text in the field of natural language generation (NLG).  Traditional n-gram-based metrics~\cite{bleu,lin-2004-rouge,snover-etal-2006-study} and embedding-based metrics~\cite{li-etal-2019-deep,zhang2020bertscore,risch-etal-2021-semantic}  can only assess lexical or semantic similarity between the generated answers and reference answers~\cite{schluter-2017-limits,reiter-2018-structured,montahaei2019jointly,freitag-etal-2020-bleu}. These metrics have been found to have a relatively low correlation with human preferences~\cite{liu2023gpteval}.  Recently, employing LLM as a judge~\cite{zheng2023judging,alpaca_eval,chan2023chateval} is a novel evaluation paradigm that has gained widespread application. The most common approach involves using proprietary LLMs, such as GPT-4~\cite{openai2024gpt4}, as judge models to rank or score outputs generated by other models. However, this method relies on closed-source models, incurs high costs, and poses risks of internal evaluation dataset leaks for companies developing LLMs. To address these issues, various works~\cite{zhu2023judgelm,li2023generative,wang2024pandalm,kim2024prometheus,kim2024prometheus2} have proposed training dedicated scoring models on open-source base models using synthetic or manually labeled data. These evaluations often use reference answers to assist in the assessment or employ general rubrics to guide the scoring process. However, these approaches overlook the differences between individual questions and the varying scoring criteria of each question, even within the same category. In contrast, we propose an evaluation paradigm based on self-adaptive rubrics that generates fine-grained, customizable rubrics for each question, guiding a more precise scoring process. It is worth noting that although Prometheus 2 also claims to use fine-grained rubrics, their rubrics remain question-agnostic.

\setlength{\parindent}{0pt}\textbf{Quantifying Evaluation Confidence.} The automatic metrics are imperfect, and we must measure their performance further. A gold standard for this is their alignment with human judgment and the confidence level we can have when these metrics guide our decision-making process. However, quantifying this performance~\cite{krishna2021hurdles, schluter2017limits,stureborg2024characterizing} is difficult due to various factors (the evaluator's accuracy and stability, evaluation set size,  the extent of the performance difference among competing models, etc.). ~\cite{kocmi2021ship,deutsch2021statistical,zhang2004measuring} investigate the correlation between human judgment and traditional automatic metrics such as ROUGE and BLEU and analyze their confidence intervals. For LLM-based evaluators, commonly used metrics include \textbf{Pearson}, \textbf{Spearman}, and \textbf{Kdendall-Tau} to measure the alignment between the model's scores and human preferences. However, previous work has primarily focused on the correlation of rankings or overall scores at the model level without comparing the scores with human ratings at the individual question. This limits the interpretability of the scoring process and hampers its utility in guiding the development and iteration of LLMs.

\section{SedarEval Benchmark}

In this section, we introduce SedarEval, a benchmark constructed upon the self-adaptive rubrics paradigm. We begin by delving into the intricacies of the self-adaptive rubric paradigm, followed by a detailed explanation of the benchmark's core components – questions and their corresponding rubrics – along with the methodology for model evaluation using this benchmark. To ensure the quality of SedarEval, we incorporate comprehensive human assessment into the construction process, meticulously filtering out samples that fail to meet the established quality standards.

\subsection{Self-Adaptive Rubrics}

Previous LLM-as-a-judge approaches, which rely on general rubrics or principles for scoring, often lack specific, problem-related rubric guidance. Consequently, these methods depend heavily on the inherent capabilities of the LLM itself, leading to potential errors in evaluations due to insufficient reasoning abilities or hallucinations. Additionally, this approach introduces extraneous biases, such as position bias and order bias.

Self-adaptive rubrics address these issues by tailoring the evaluation criteria to the specific problems at hand, incorporating the focal points of the problem and assigning different weights accordingly. By introducing penalty points, these rubrics align more closely with human judgments by deducting points for outputs that deviate from expected tendencies. To prevent human evaluators (or LLMs) from making incorrect assessments due to a lack of background information, additional context is provided for each question to assist in the scoring process. A typical self-adaptive rubric comprises three components: scoring points, penalty points, and background knowledge, as illustrated in Table 3.

\subsection{Dataset Construction}

\textbf{Questions:} We have defined a classification system for objective questions, with a two-tiered scoring system as shown in the diagram. Under each secondary classification, we have hired five people to create questions. Specifically, each person is required to first create their own questions to get a question pool, and then each person votes on all the questions. We only keep the questions that all five people agree on.

For each candidate question, the annotators will select 5 LLMs to test the effectiveness of the questioned question.
We only keep the questions with a larger variance in scores, which are more discriminating, and remove the questions where the answers from different models are almost the same, which are not helpful in distinguishing between different models. For example, if a question can be answered correctly by all models, or incorrectly by all models, then this question cannot show which model is better.

After collecting the initial questions, we hired another group of people to compare all the questions in pairs to judge the similarity of the problem-solving ideas for the two questions and delete the questions with too much similarity.

\textbf{Rubrics:} For each question, we assign it to three individuals to discuss together and generate a rubric like the one shown in Figure~\ref{fig:pipeline}.

For more detailed information, please refer to Appendix~\ref{app:benchmark}, which contains benchmark statistics and the leaderboard.
\subsection{Evaluation Pipeline}

The entire evaluation pipeline using our benchmark is illustrated in Figure~\ref{fig:pipeline}. Given a question, its corresponding rubrics, and the model to be evaluated, we first input the question into the model to generate a response. The response is then scored according to the predefined rubric, either by human evaluators or using LLMs. Finally, all the scores are aggregated to obtain the model's total score.

\section{Evaluator Language Model}

In this section, we introduce an evaluator LM aligned with the self-adaptive rubrics paradigm to substitute human evaluators. We begin by delineating the evaluation format. Subsequently, we propose a novel data filtering strategy to align the Chain-of-Thought evaluation process with human judgments. Finally, we discuss the automation of rubric generation.

\subsection{Evaluation Format}

The evaluation format consists of two types: direct scoring of individual model outputs and pairwise comparison of model outputs to determine the superior one. Pairwise evaluation requires significantly more comparisons as the number of candidate models increases, as shown by Equation~\ref{eq: direct}. Therefore, we employ direct assessment in this paper. Notably, direct assessment scores can be compared to derive pairwise results.

\begin{equation}
\label{eq: direct}
C(n, 2) = \frac{n!}{2!(n-2)!} - n = \frac{n^2 - 3n}{2}
\end{equation}

We use a reference-based format to organize the output.  Specifically, for each question, we compile the reference answer, self-adaptive rubrics, and scoring examples to create an auto-prompt template. When evaluating answers, we incorporate the answers into this auto-prompt template as the complete input. We conduct ablation experiments on each component in zero-shot, few-shot, and instruction tuning settings.

\subsection{Human-AI Consistency}

Human annotators provide specific scores for each response without corresponding explanations, which is efficient but suboptimal for training evaluator LMs. To alleviate this issue, we use GPT-4 to generate detailed reasoning steps using Chain-of-Thought. However, scoring preferences may differ between GPT-4 and human annotators, and both may make errors. To mitigate these errors and align the scoring process with human judgment, we introduce a \textbf{Human-AI Consistency} strategy to improve synthetic data quality. We extract final scores from the GPT-4 scoring process and compare them with human scores, retaining only the data where GPT-4 and human results are consistent, as shown in Equation~\ref{eq:T}, where $\mathcal{H}$ represents human scores, $\mathcal{A}$ represents AI scores, and $\mathbb{I}$ is an indicator function.

\begin{equation}
\label{eq:T}
\mathcal{T} = \left\{ (h, a) \mid h \in \mathcal{H}, a \in \mathcal{A}, \mathbb{I}(h, a) = 1 \right\}
\end{equation}

This approach only retains instances where human and AI scores are consistent and differs from rejection sampling, which uses human scores as a reward function to select the optimal output from multiple GPT-4 results.

\subsection{Automatic Rubric Generation}

To reduce human annotation costs, we investigate using human-annotated datasets to train a model for automatic self-adaptive rubric generation. By providing the model with questions and corresponding reference answers, we train it to produce rubrics that delineate scoring criteria and identify deduction points.

Generating self-adaptive rubric format output is straightforward, but aligning rubrics with human preferences requires aligning the model with human evaluative criteria. This complexity arises because identifying scoring points, assigning specific weights, and criteria for deductions are significantly influenced by human judgment.

The training process for the automatic rubric generation model comprises two stages. Initially, we use human-labeled data to train a base model through Supervised Fine-Tuning (SFT), as depicted in Equation~\ref{eq:SFT}.

\begin{equation}
\label{eq:SFT}
\mathcal{L}(\theta) = - \sum_{i=1}^N \log p_{\theta}(y_i | x_i)
\end{equation}

The base model generates rubrics that conform to the specified format, though they may not fully align with human preferences (quantitative metrics will be introduced in Section~\ref{sec:metrics}). In the next phase, rubrics generated by the base model are treated as rejected responses, denoted as $y_{l}$, while human-labeled rubrics serve as preferred responses, denoted as $y_{w}$, to construct preference pairs. We then train the model using Direct Preference Optimization (DPO)~\cite{rafailov2024direct} to align it with human preferences, as shown in Equations~\ref{eq:dpo_1} and~\ref{eq:dpo_2}.

\begin{equation}
\label{eq:dpo_1}
f(y, x) = \beta \log \frac{\pi_{\theta}(y \mid x)}{\pi_{\text{ref}}(y \mid x)}
\end{equation}

\begin{equation}
\label{eq:dpo_2}
\mathcal{L}(\pi_{\theta}) = - \log \sigma \left( f(y_w, x) - f(y_l, x) \right) \quad
\end{equation}

We also explore automating rubric generation using GPT-4 without reference answers. To ensure accuracy, GPT-4 creates both the rubric and an ideal answer for each question. If the ideal answer corresponds with the ground truth, the generated rubric is deemed acceptable. We employ a self-refinement strategy to help the model iteratively refine its outputs, aligning it with human preferences. For detailed algorithmic procedures and prompts, refer to Appendix~\ref{app:rubric}.



\section{Experiment}
\subsection{Experimental Setting}
We train the Evaluator Language Model and the Rubric Generation Model using both the open-source model LLaMA-3~\citep{touvron2023llama} and our internal model XD\footnote{The name of this model has been anonymized to ensure confidentiality.}. To maximize training efficiency and utilize hardware resources, we implement tensor parallelism~\citep{shoeybi2020megatronlm} with PyTorch 2.3~\citep{paszke2019pytorch}. For the 7B/8B models, we use 128 H100 GPUs, while for the 70B models, we use 512 H100 GPUs. For the models' chat versions (i.e., instruction-tuned), we employ the same chat markup language (ChatML) as the models themselves. For the pre-trained versions, we use a unified ChatML to reduce data bias. We adopt adaptive learning rate and batch size strategies. Further training details are provided in Appendix~\ref{app:train}.

\subsection{Evaluation Metrics}
\label{sec:metrics}
To assess the performance of the evaluator language model, we use Pearson's correlation coefficient and Spearman's rank correlation coefficient. These statistical measures assess the consistency between the outcomes of the evaluator language model and those obtained from human evaluators.

Each question is accompanied by a detailed rubric specifying exact scoring and deduction criteria, so we use accuracy to evaluate the model's capability in following these self-adaptive rubrics for scoring. Considering potential noise in the model scoring, we introduce a weaker threshold ACC, which considers a result correct if it falls within a specified range. The calculation formulas are presented in Equation~\ref{eq:acct}.

\begin{equation}
\label{eq:acct}
\text{ACC}_t = \frac{1}{N} \sum_{i=1}^{N} \begin{cases} 
1, & \text{if } |y_{\text{pred}_i} - y_{\text{true}_i}| \leq \epsilon \\
0, & \text{otherwise}
\end{cases}
\end{equation}

To facilitate the iterative enhancement of LLMs using our benchmark, a robust metric is essential to assess whether the current model version outperforms its predecessor. Therefore, we adopt the widely used GSB (Good, Same, Bad) metric to compare model performance. Given two models, A and B, the calculation formula is presented in Equation~\ref{eq:gsb}. In this context, "\#good" signifies that model A surpasses model B, "\#bad" denotes the contrary, and "\#same" indicates equivalent performance between the models.

\begin{equation}
\label{eq:gsb}
\Delta GSB = \frac{\# \text{good} - \# \text{bad}}{\# \text{good} + \# \text{same} + \# \text{bad}}
\end{equation}

To evaluate the quality of automatically generated rubrics, we draw on the ACU~\citep{liu-etal-2023-revisiting} and FactScore~\citep{min-etal-2023-factscore} paradigms, using GPT-4 to calculate the match between the generated rubrics and the ground truth rubrics. The calculation formula is specified in Equation~\ref{eq:match}, where GT represents the correct rubric set containing multiple \{grading points: specific score\} pairs, and AT denotes the automatically generated rubric set. \(\mathbb{I}(i \in GT)\) is an indicator function that equals 1 if the item \(i\) from AT is present in GT, and 0 otherwise. The prompts used for this evaluation are detailed in Appendix~\ref{app:set}.

\begin{equation}
\label{eq:match}
\text{Match}(GT, AT) = \frac{\sum_{i \in AT} \mathbb{I}(i \in GT)}{|GT|}
\end{equation}

\subsection{Selected Models}
Previous studies predominantly employ English-proficient models to generate \textless question, response, score\textgreater{} triples for training evaluator language models, often overlooking models proficient in Chinese. Additionally, several studies exclusively use GPT-3.5 or GPT-4 to construct such synthetic data. These data generation methodologies may cause discrepancies between the synthetic and real-world data distributions, introducing biases into the trained evaluator language models.

To alleviate this issue, we utilize a broader range of LLMs to collect responses that better reflect real-world distributions. This approach ensures greater diversity and mitigates biases introduced by relying solely on synthetic data generated from a single model. Specifically, we choose GPT-4, GPT-4-turbo, GPT-4-o, Claude Opus, DeepSeek 2.0, MiniMax 6.5, MiniMax 6, Doubao, GPT-3.5, Tongyi Qianwen 2.0, and Tongyi Qianwen 1.5-100B/70B. This selection includes models proficient in different languages and multiple versions of the same model.

For open-source models, we use local deployment to infer responses. For proprietary LLMs, if API services are available, we collect model outputs by requesting the API. If only a web interface is provided, we employ people to gather the outputs.

\section{Analysis}

In this section, we conduct a comprehensive experimental analysis of the robustness of the proposed benchmark evaluations, examining the data distribution, training phases, and training paradigms of evaluator LMs. Our findings reveal limitations in current training methodologies for evaluator LMs. Building on these insights, we develop an evaluator LM aligned with the self-adaptive rubrics paradigm.

\subsection{Scaling Law for Robust Evaluation}
A robust benchmark should effectively distinguish the capabilities of different models and maintain stability to ensure consistent rankings rather than allowing fluctuations due to the instability of individual tasks. To achieve this, the benchmark needs a sufficiently broad distribution while minimizing extraneous biases.

To verify the robustness of the proposed benchmark, we conduct two rounds of sampling without replacement from a pool of 1,000 questions. In each round, we select \( n \) questions, resulting in a total of \( 2n \) independent questions, where \( n \in [10, 500] \). We then compare the consistency of the model rankings obtained from these two samples.

\begin{figure}[h]
    \centering
    \includegraphics[width=\linewidth]{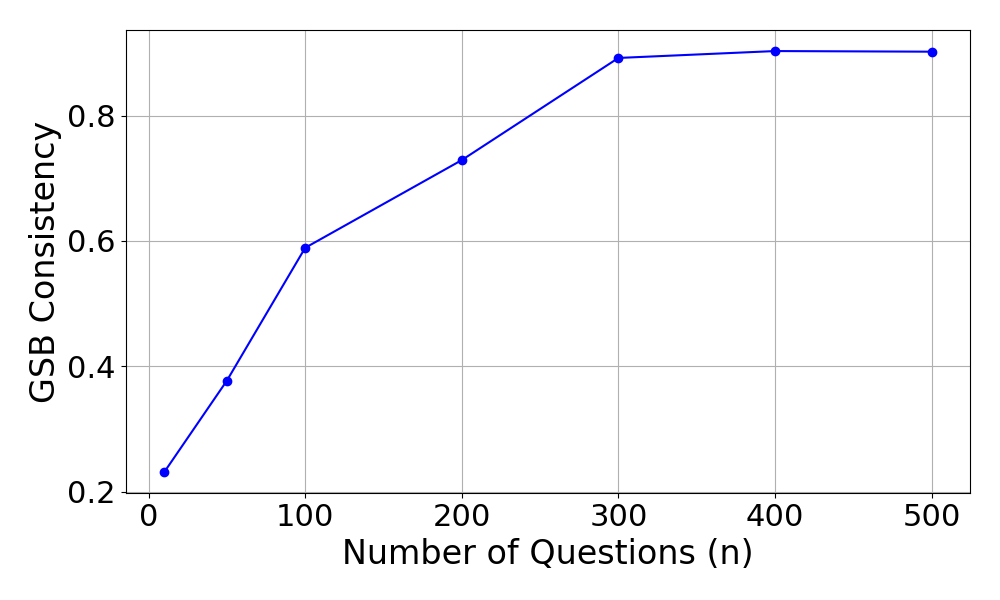}
    \caption{Consistency of model rankings as \( n \) increases. After \( n \) reaches approximately 300, the consistency stabilizes with only minor fluctuations.}
    \label{fig:consistency}
\end{figure}

Figure \ref{fig:consistency} shows the variation in the consistency of model rankings under different question sets as \( n \) increases. When \( n \) is relatively small, the consistency is low, indicating the inconsistency caused by biases in different distributions. As \( n \) increases, the consistency improves despite the two sets remaining independent. After \( n \) reaches approximately 300, the consistency stabilizes with only minor fluctuations. This demonstrates the scaling law for robust evaluation, indicating that as the number of questions increases, the evaluation results tend to stabilize due to the broader coverage of the distribution.

\subsection{Score Distribution Shift}

\begin{figure}[h]
    \centering
    \includegraphics[width=\linewidth]{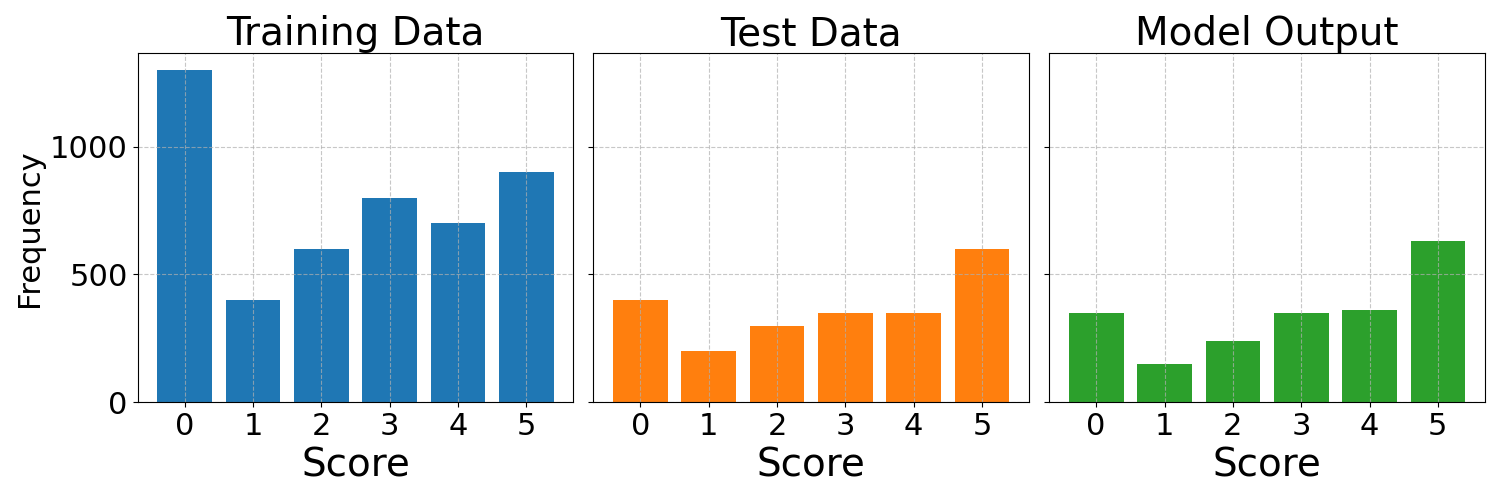}
    \caption{Data distribution comparison of different data.}
    \label{fig:distribution}
\end{figure}

The Prometheus approach, which relies on general rubrics not specifically tailored to the problem at hand, employs GPT-4 to generate an equal amount of synthetic data across different scores (1-5) to mitigate score bias from the evaluating language model. In contrast, our method uses self-adaptive rubrics, and our responses are genuinely collected from the model rather than artificially synthesized. Consequently, we cannot ensure that the quantity of data for each score is perfectly balanced.

However, as illustrated in Figure~\ref{fig:distribution}, we observe that despite the score distribution shift between the training and test data, the score distribution of the model outputs, trained using the self-adaptive rubrics paradigm, closely aligns with the human-provided ground truth. This finding substantiates the robustness and efficacy of the self-adaptive rubrics paradigm in automated scoring.

\subsection{Out of Distribution Evaluation}
We establish two dimensions for evaluating the out-of-distribution capabilities of our model: model-level and question-level. For the model-level evaluation, we utilize the same set of questions, selecting a subset of models to train the evaluator language model (LM), and subsequently test on the remaining unselected models. In the question-level evaluation, a subset of questions along with all associated models are used for training, and the scoring performance is then assessed on a different set of questions.

\begin{table*}[ht]
\centering
\begin{tabular}{lcccccccc}
\hline
& \multicolumn{4}{c}{Question-level} & \multicolumn{4}{c}{Model-level} \\
\cline{2-5} \cline{6-9}
Type & GSB & ACC & ACC(t) & pearson & GSB & ACC & ACC(t) & pearson \\
\hline
XD & 0.952 & 0.590 & 0.794 & 0.738 & 0.952 & 0.590 & 0.794 & 0.380 \\
GPT-3.5 & 0.829 & 0.422 & 0.663 & 0.566 & 0.829 & 0.422 & 0.663 & 0.566 \\
GPT-4 & 0.952 & 0.654 & 0.855 & 0.822 & 0.952 & 0.654 & 0.855 & 0.822 \\
\hline
\end{tabular}
\caption{Out of distribution evaluation performance in both model-level and question-level.}
\label{tab:out_of_distribution}
\end{table*}

Table~\ref{tab:out_of_distribution} presents the experimental results, showing that under the self-adaptive rubrics paradigm, the model performs well in both model-level and question-level evaluations. This indicates that our proposed method has strong generalization capabilities.

\subsection{Merged SFT or Continual SFT}

Previous research shows that a model might generate a correct answer but fail to accurately evaluate the <question, answer> pair for the same question. We argue that this issue mainly arises from the insufficient diversity of the SFT data.

To validate this, we conduct the following experiments:
\begin{enumerate}
\item Training a pretrained language model (PLM) using only traditional SFT data.
\item Training a PLM using a mix of SFT data and evaluator LM format data.
\item Performing continual SFT on an instruction-tuned model using evaluator LM format data, a widely adopted approach in other studies~\cite{kim2024prometheus2, kim2024prometheus}.
\end{enumerate}

\begin{table}[ht]
  \centering
  \caption{Experiments on training phases and training data, where v1 represents continual SFT, v2 represents SFT from PLM, v3 represents SFT incorporating evaluator LM format data, and v4 represents data filtered using the Human-AI Consistency strategy.}
  \label{tab:hac}
    \setlength{\tabcolsep}{0.6mm}{
  \begin{tabular}{lccccc}
    \hline
    Type & GSB & ACC & ACC(t) & pearson & general \\
    \hline
baseline & 0.784 & 0.339 & 0.584 & 0.263 & 730 \\
XD-v1 & 0.910 & 0.514 & 0.755 & 0.686 & 458 \\
XD-v2 & 0.895 & 0.551 & 0.802 & 0.761 & 684 \\
XD-v3 & 0.911 & 0.593 & 0.811 & 0.765 & 653 \\
XD-v4 & \textbf{0.941} & \textbf{0.664} & \textbf{0.854} & \textbf{0.829} & 685 \\
     \hline
  \end{tabular}
  }
\end{table}

As shown in Table~\ref{tab:hac}, we find that although the model using continual SFT performs well on evaluation tasks, its general ability significantly declines, limiting its versatility. However, starting from a PLM and using a mix of SFT data and evaluator LM format data for SFT results in excellent evaluation capability with minimal impact on general ability. This reveals the shortcomings of the previous continual SFT approach and suggests that the model's inability to evaluate the questions it can answer may simply be due to the lack of such data, highlighting the importance of diversity in SFT data.

We employ Human-AI Consistency to filter the evaluator LM and find that, compared to using raw Chain-of-Thought data generated by GPT-4 and data filtered by rejection sampling, the data selected using Human-AI Consistency shows significant improvements in both evaluation and general capability, demonstrating the effectiveness of this strategy.

\subsection{Ablation Study}
We conduct detailed ablation experiments on the components of self-adaptive rubrics, namely, reference answers, rubrics, and in-context examples. As shown in Table~\ref{tab:alb_com}, the consistency between the evaluator LM and human scoring significantly increases after incorporating self-adaptive rubrics. However, the improvements are not as pronounced when other components are added, indicating that the primary driver of enhanced performance is the self-adaptive rubrics themselves. This suggests that self-adaptive rubrics play a crucial role in aligning the evaluator LM with human judgment.

\begin{table}[ht]
\centering
\begin{tabular}{lcccc}
\hline
Type & GSB & ACC & ACC\_t & Pearson \\
\hline
Baseline & 0.963 & 0.636 & 0.802 & 0.733 \\
+ rubric & 0.957 & 0.706 & 0.871 & 0.843 \\
+ R.A & 0.952 & 0.717 & 0.877 & 0.848 \\
+ example & 0.959 & 0.728 & 0.888 & 0.867 \\
\hline
\end{tabular}
\caption{Ablation study for each component, where R.A. stands for reference answer.}
\label{tab:alb_com}
\end{table}

\subsection{Comparison with Alternative Paradigm}
Using the same training data, we conduct a comparative analysis between the self-adaptive rubrics paradigm and the existing general rubric paradigm, as presented in Table~\ref{tab:ablation_01}. The results demonstrate that our approach significantly outperforms existing methods. Furthermore, in addition to accurately ranking the models, our method provides fine-grained capability evaluations that closely align with human assessments. This is both crucial and practical for facilitating the iterative development of LLMs. Due to space constraints, detailed descriptions and results of other experiments are provided in Appendix~\ref{app:additional_experiments}.

\section{Conclusion}

In this paper, we introduce a novel evaluation paradigm called self-adaptive rubrics, aligning the scoring process with human judgment and reducing bias by tailoring rubrics to specific questions. Based on this paradigm, we develop a new benchmark, INSDA. To automate scoring, we analyze existing open-source evaluator language models and identify training phase data diversity issues. We then introduce human-AI consistency to align the chain-of-thought evaluation with human judgment and propose an evaluator LM that follows the self-adaptive rubrics paradigm. Experimental results show our model achieves higher consistency with human evaluation compared to GPT-4. We hope our work inspires researchers to apply this paradigm to more tasks, aligning automated scoring with human judgment.

\section*{Limitations}

In this paper, we propose an evaluation paradigm based on self-adaptive rubrics, which provides more granular process guidance to align the scoring process with human judgment. Additionally, we introduce a benchmark, INSDA, based on this framework. However, there are several limitations:
\begin{itemize}
    \item For questions with multiple correct answers, it requires manually writing multiple self-adaptive rubrics. It is worth noting that, to our knowledge, no current work focuses on the multi-solution direction.
    \item For subjective questions, such as creative writing, poetry, and other forms of artistic expression, different groups or individuals may have varying definitions of what constitutes good work. Therefore, it is necessary for each group or individual to set their own self-adaptive rubrics rather than relying on predefined ones. This also highlights the flexibility and interpretability of the self-adaptive rubrics paradigm we propose.
\end{itemize}

\section*{Ethical Considerations}

We propose a scoring paradigm based on self-adaptive rubrics to enhance the interpretability and controllability of the automated scoring process. This approach aims to improve the credibility of evaluation results produced by LLMs and to support the research community in advancing these models. Nevertheless, the inherent hallucinations within LLMs pose a challenge to ensuring the complete accuracy of automated evaluation outcomes. Therefore, we recommend incorporating human review of certain outputs when using LLMs as judges to increase the overall reliability and credibility of the process.

Additionally, when generating self-adaptive rubrics for subjective questions, different groups or individuals may have varying definitions of what constitutes a good answer, potentially leading to biases and discrepancies. We encourage dialogue and mutual understanding among groups or individuals with diverse values, promoting the use of self-adaptive rubrics that align with their respective values and preferences.



\bibliography{custom}
\newpage
\appendix

\section{Training Details}
\label{app:train}
We employed a default learning rate of 2e-5, and the batch size per device was dynamically adjusted based on the total data volume and the number of machines to maintain consistent optimization steps. The Adam optimizer was utilized. For the scaling law experiments, when \( n \) was below 300, we performed three repetitions and averaged the results to minimize error. For all invoked APIs, we used the default parameters without extensive modifications.

\section{Data Annotation}
\label{app:data}
\subsection{Annotator Qualifications}
\begin{figure*}[t]
  \includegraphics[width=0.9\linewidth]{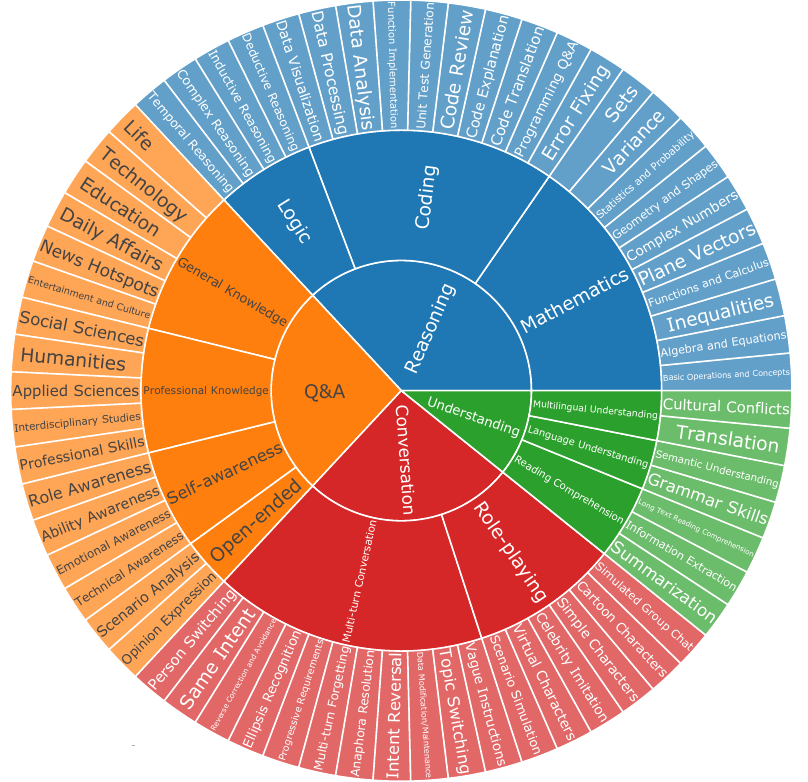}
  \caption {Benchmark multi-level classification system. To maintain conciseness, we have detailed the hierarchy only up to the third tier.}
  \label{fig:tax}
\end{figure*}

All our annotators are internal team members with at least a Master's degree. We provide additional compensation significantly higher than the standard salary, based on the amount of data annotated. The taxonomy of Sedareval dataset is shown in Fig~\ref{fig:tax}.

\section{Prompt Templates}
\label{app:tem}

\subsection{AutoPrompt}
\label{app:auto}

\begin{myquote}
Below, I will provide a \textless Question\textgreater, along with the corresponding \textless Reference Answer\textgreater\ and \textless Scoring Rubric\textgreater. You need to evaluate the \textless Output Result\textgreater\ from the \textless Model Answer\textgreater\ of the \textless Model to be Assessed\textgreater. The evaluation should be divided into two parts: "Scoring Process" and "Final Score." Please note that the scoring range is from 0 to 5 points. You must justify the score you assign based on the \textless Model Answer\textgreater, strictly adhering to the requirements of the \textless Scoring Rubric\textgreater\ without adding, changing, or imagining any additional criteria.

\end{myquote}

\subsection{Prompt for Set Matching}
\label{app:set}

\begin{myquote}
You are a meticulous judge tasked with evaluating whether the "Test Rubric" provided by the user aligns with the "Standard Rubric." The evaluation rules are as follows:
\begin{itemize}
    \item The initial total score is set to zero.
    \item For each item in the "Test Rubric":
    \begin{enumerate}
        \item If the item matches any item in the "Standard Rubric" exactly, one point is added to the total score.
        \item If the item in the "Test Rubric" is unrelated to any item in the "Standard Rubric," the total score remains unchanged.
        \item If the item in the "Test Rubric" is the exact opposite of any item in the "Standard Rubric," one point is subtracted from the total score.
    \end{enumerate}
\end{itemize}

You need to return the entire scoring process (explaining why points were added or subtracted) along with the final score. The return format should be:

\begin{quote}
\{
    "Scoring Process": "\textless Here, provide the scoring process as a string\textgreater",\\
    "Final Score": "\textless Here, provide the final score as a mathematical expression, concluding with 'Final Score: \textless score\textgreater' e.g., '3/5=0.6, Final Score: 0.6'\textgreater"
\}
\end{quote}

The returned format must be compatible with \texttt{json.loads()} to be converted into a dictionary.
\end{myquote}

\subsection{Prosecutor Prompt}
\begin{myquote}
Please check if the generated answer is correct. The reference answer is: \{gt\}, and the generated answer is: \{user\}.
Please respond in the following format:
\{
    "result": True
\}
\end{myquote}

\subsection{Refinement Prompt}
\begin{myquote}
Your generated answer is not the standard answer. Please reflect on this and generate a new answer. 

The generated scoring points and the full score answer are:
\end{myquote}

\begin{figure*}[h]
  \includegraphics[width=0.9\linewidth]{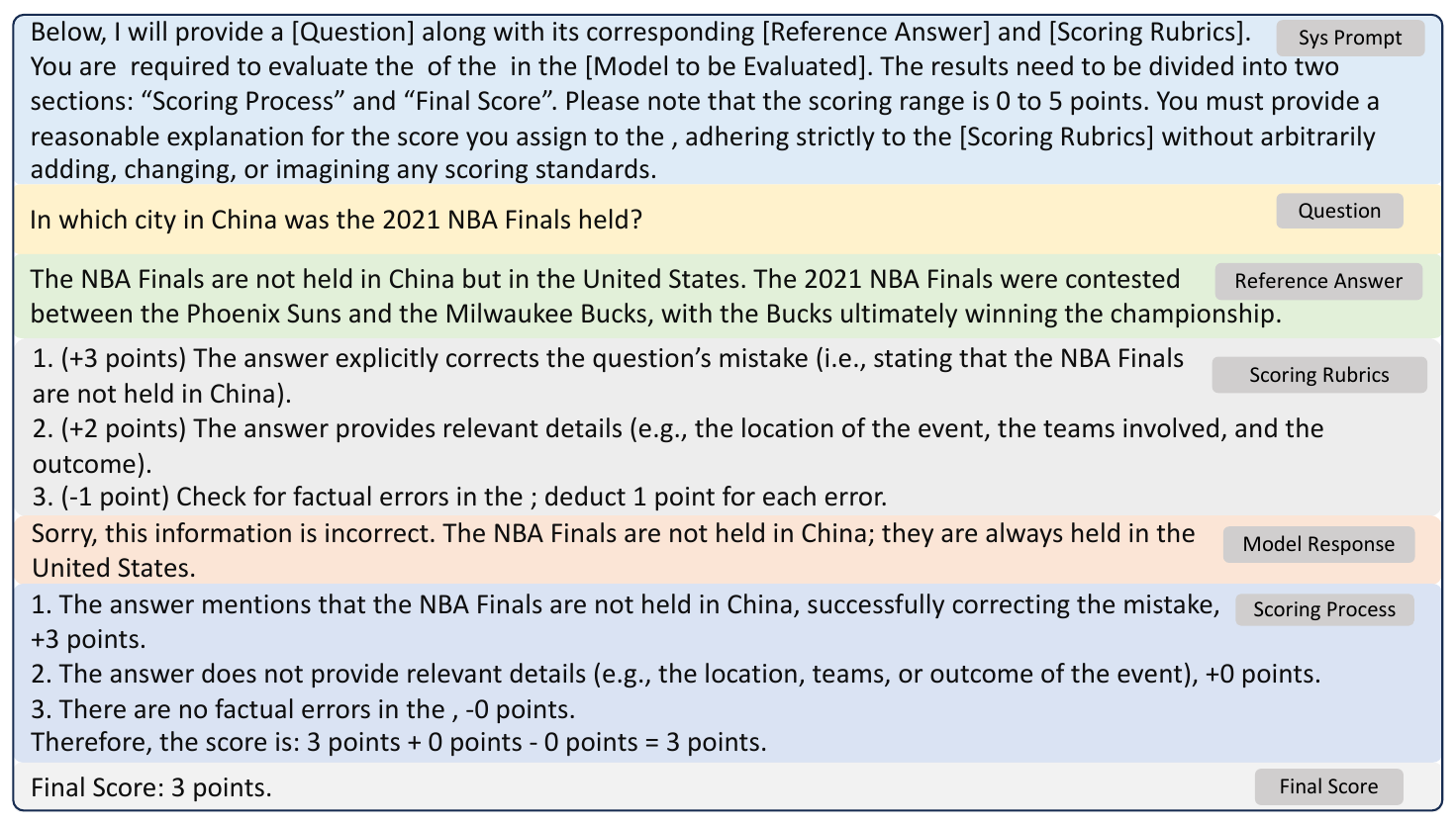}
  \caption { An illustrative example of the format used in self-adaptive rubrics.}
\end{figure*}




\section{Additional Experiments}
\label{app:additional_experiments}

\subsection{Automatic Rubric Generation}
\label{app:rubric}

In this section, we conduct experiments on automatic rubric generation and analyze the results to explore its potential. Specifically, we compare the following three experimental setups: 1) using GPT-4 to automatically generate rubrics; 2) training an automatic rubric generation model with human-labeled rubrics under the SFT paradigm; and 3) aligning the model output with human preferences using DPO.

\begin{algorithm}
\caption{Self-Adaptive Rubrics Generation and Validation}
\label{alg:rubrics}
\begin{algorithmic}[1]
\Require $Q$ \Comment{Given question}
\Require $GT$ \Comment{Ground truth answer}
\Require $n$ \Comment{Maximum iterations}
\State $i \gets 0$
\State $accepted \gets \text{False}$
\While{$i < n$ \textbf{and} $\neg accepted$}
    \State $R, IA \gets \text{GPT-4}(Q)$ \Comment{Generate rubrics and ideal answer}
    \If{$PA(IA, GT)$} \Comment{Prosecutor agent checks ideal answer}
        \State $accepted \gets \text{True}$
    \Else
        \State Inform GPT-4 of incorrect $IA$
        \State $i \gets i + 1$
    \EndIf
\EndWhile
\If{$accepted$} \State
    \Return $R$
\Else \State
    \Return \text{Failure in } $n$ \text{ iterations}
\EndIf
\end{algorithmic}
\end{algorithm}


\subsection{Error Propagating}


When using rubric generation models to automatically create self-adaptive rubrics, a potential issue is that if the generated rubric is inconsistent with the human-provided rubric, errors can accumulate in the scoring pipeline, leading to a larger deviation in the final score. By incorporating a filtering strategy, the overall performance will improve.





\subsection{Joint Training vs. Expert Training}

We also explored whether to combine data from different categories for joint training when training the evaluator LM or rubric generation model, or to train a separate expert model for each category individually. We found that using joint training can achieve better results than expert training.

\end{document}